\documentclass[letterpaper, 10 pt, conference]{ieeeconf}  

\IEEEoverridecommandlockouts                              
\usepackage{graphicx} 
\usepackage{amsmath} 
\usepackage{subfig}
\usepackage{multirow}
\usepackage{algorithm}
\usepackage[noend]{algpseudocode}
\usepackage[obeyspaces]{url}
\makeatletter
\def\BState{\State\hskip-\ALG@thistlm}
\makeatother

\title{Fast Gaussian Process Occupancy Maps
}

\author{Yijun Yuan, Haofei Kuang and S\"oren Schwertfeger
\thanks{The authors are with the School of Information Science and Technology, ShanghaiTech University, Shanghai, China
        {\tt\small [yuanwj, kuanghf, soerensch]@shanghaitech.edu.cn}}%
}

\begin{document}

\maketitle
\thispagestyle{empty}
\pagestyle{empty}

\begin{abstract}
In this paper, we demonstrate our work on Gaussian Process Occupancy Mapping (GPOM). We concentrate on the inefficiency of the frame computation of the classical GPOM approaches. In robotics, most of the algorithms are required to run in real time. However, the high cost of computation makes the classical GPOM less useful. In this paper we don’t try to optimize the Gaussian Process itself, instead, we focus on the application. By analyzing the time cost of each step of the algorithm, we find a way that to reduce the cost while maintaining a good performance compared to the general GPOM framework. From our experiments, we can find that our model enables GPOM to run online and achieve a relatively better quality than the classical GPOM.
\end{abstract}

\section{INTRODUCTION}
\label{sec:1}

Mapping is an important research area in the field of robotics. Occupancy grid mapping has been introduced already about 3 decades ago and it has been widely used in the field of robotics for its simplicity and computation efficiency. It is quite often at the heart of a Simultaneous Localization and Mapping (SLAM) system, where it integrates the sensor data (most often (laster) range data) into a map. Occupancy grid mapping is simple in assuming that, upon updating the map from a new scan, the probability of occupation of a cell is solely determined by the ray passing through. This is loosing the spatial context information, because in reality the occupancy of a cell can often be related to its neighboring cells \cite{c1}.

Gaussian Process Occupancy Mapping (GPOM) is a mapping algorithm which is updating all relevant cells when integrating a scan. GPOM has been developed for about a decade. It has its advantage over general occupancy mapping and some other probability-based maps to better model the sparsely sampled points and its fit for structural information. In \cite{c1} the authors build the whole map at a time, which takes quite a long time, because holding all of the range sensor data in one formula takes too much computation. What’s more, the algorithm cannot incrementally integrate new data, which is limiting the utility of GPOM.

Inspired by Bayesian Committee Machine (BCM), \cite{c3} clusters laser scans into groups. Instead of building the whole map in one go, as \cite{c1}, they separate the task into many subproblems. By building the map on each group and fusing them with BCM, the computation cost of each group is tremendously reduced and, more importantly, this approach makes the algorithm incremental. As an improvement, \cite{c6} and \cite{c7} incrementally update the map in each frame, which allows for exploring in dynamically expanding environments. 

After that, a more scalable method, Hilbert Maps \cite{c4}, has been proposed. Comparing with GPOM, the Hilbert Maps method is a parametric method that can be trained with Stochastic gradient descent (SGD). However, it is less accurate than GPOM and, more importantly, it has to pass all the data multiple times and does not run incrementally.

Inspired by the indoor mapping survey \cite{c2}, many indoor mapping algorithms are incremental such that it is possible to run them in real time. The authors of this paper were wondering why GPOM has not been applied in real time, even though it can run incrementally already. The answer is, that, even for the current incremental GPOM algorithms, the computing requirements are still intractable.

The Gaussian Process regression (GP) is the source of restrictions. Generally, the cost to build the model is $\mathcal{O}(n^3)$, while the prediction only takes $\mathcal{O}(n)$. While GPOM also requires variance, the prediction of GPOM will take $\mathcal{O}(n^2)$.

Most of the work to optimize GP is done on its training process. But the speed limit may not be the same in GPOM application. In order to find the bottleneck that makes GPOM not meet the needs of real-time processing, we analyzed the GPOM pipeline as follows:

\begin{enumerate}
	\item	Extract positive and negative samples
	\item Set the testing sample
	\item	GP model Building
	\item	Prediction on the test set to achieve $\mu$ and $\sigma$ (see Section \ref{sec:2})
	\item	Fuse $\mu$ and $\sigma$ of the global frame (size $W_G\times H_G$) with the local frame (size $W_l\times H_l$)
	\item	Build the image with logistic regression function
	\item	Publish with ROS
\end{enumerate}


In our experiments we found, that most of the computation cost is not in the third step (model building). This is a surprising result which is contradicting most of the works on GP. Instead, we found that the cost in prediction, step 4, is the highest.

In this paper, we are thus proposing a GPOM algorithm that optimizes the prediction step, which then makes our Fast-GPOM can run at least 10 times faster than GPOM in that step.

\begin{figure}[tpb]
	\centering
	\subfloat[Fast-GPOM]{
		\label{fig:yellowf}
		\includegraphics[width=0.45\linewidth]{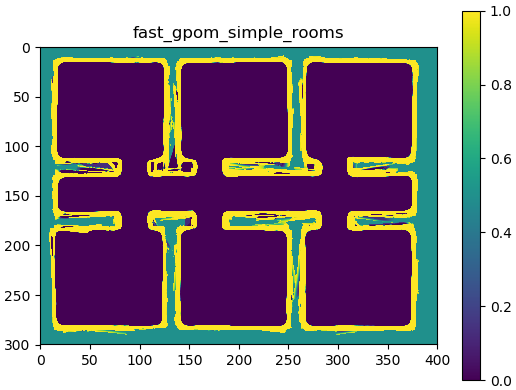}}
	\subfloat[GPOM]{
		\label{fig:yellogpom}
		\includegraphics[width=0.45\linewidth]{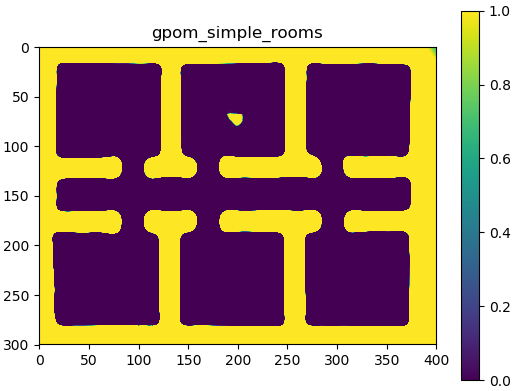}}\\
	\subfloat[Ground Truth]{
		\label{fig:yellogt}
		\includegraphics[width=0.45\linewidth]{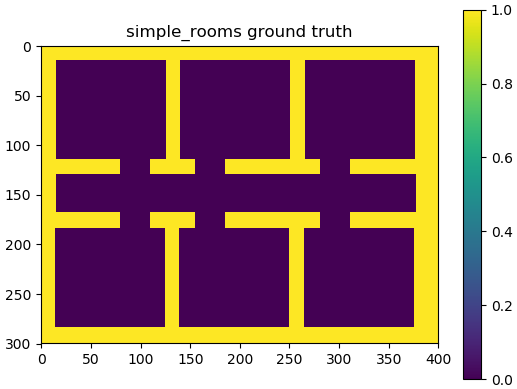}}
	\subfloat[ROC]{
		\label{fig:yelloauc}
		\includegraphics[width=0.45\linewidth]{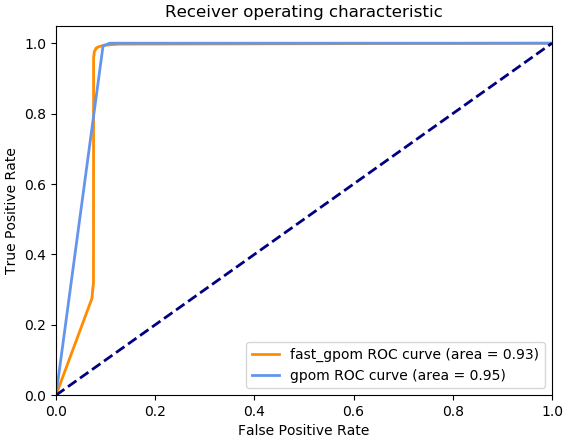}}
	\label{fig:yellow}
	\caption{Our Fast-GPOM and GPOM on a synthetic map. The color bar shows the probability to be occupied.}
\end{figure}

Additionally, we observed that GPOM has a design defect. In Fig. \ref{fig:yellogpom} a wide range of space outside GPOM has high a probability of beginning occupied. Out Fast-GPOM can mitigate such problems.

This paper is organized as follows: In the Section \ref{sec:2}, we review the GP and GPOM algorithm to make the further analysis clear. In Section \ref{sec:3}, by concentrating on the bottleneck, we propose our improved algorithm. After that, Section \ref{sec:4} presents experiments to show the properties of our method. We conclude our work in Section \ref{sec:5}.

\section{Background}
\label{sec:2}
\subsection{Gaussian Process}

Gaussian Processes (GP) generally have been applied to many tasks, but mainly for regression and classification. By comparing the similarity between features of observation and the inference sample, GP can provide both estimation and uncertainty in one step.

With pairs $\{ X_n,y_n  \}\mid _{n=1,\cdots,N}$, the regression function $f$ to solve here in GP is 
\begin{align}
y_n = f(X_n) + \epsilon\text{,  }\epsilon\sim \mathcal{N}(0,\sigma^2)
\end{align}

The latent function in the above formula is defined as 
\begin{align}
f(X) = w^T \phi (X)
\end{align}

We can then obtain 
\begin{align}
f(X^*) = \mathcal{N}(\mu, \sigma)
\end{align} with the assumption that the target data is jointly Gaussian.

%


With the kernel function $K(X_1,X_2) = \phi (X_1)^T\Sigma \phi (X_2)$, the predicted mean $\mu$ and standard derivation $\sigma$ in the above formula should be:
\begin{align}
\label{eq:mu}
\mu^* = K(X^*,X)(K(X,X)+\sigma_n^2 I)^{-1}y
\end{align}
\begin{align}
\label{eq:sigma}
\sigma^* = K(X^*,X^*) - K(X^*,X)(K(X,X)+\sigma^2 I)^{-1}K(X,X^*)
\end{align}
\begin{algorithm}
	\caption{GP Regression}\label{algo:GPR}
	\begin{algorithmic}[1]
		\Require $(\mathbf X,\mathbf y)$, $X^*$, $\sigma^2_n$
		\State $\mathbf L \leftarrow cholesky(\mathbf K(\mathbf X,\mathbf X)+\sigma^2_n\mathbf I)$
		\State $\mathbf \alpha \leftarrow \mathbf L^T \setminus (\mathbf L \setminus \mathbf y)$
		\State $\mu^* \leftarrow K( X^*,\mathbf X)\alpha$
		\State $\mathbf v \leftarrow \mathbf L \setminus K(\mathbf X, X^*)$
		\State $\sigma^* \leftarrow K( X^*, X^*)-\mathbf v^T\mathbf v$
		\State	\Return $\mu^*$, $\sigma^*$
	\end{algorithmic}
\end{algorithm}

To make it more clear how the computation is separated in training and inferencing, we follow \cite{c8} to write the algorithm in Algorithm \ref{algo:GPR}. In Algorithm \ref{algo:GPR}, $(\mathbf X,\mathbf y)$ are the observed pairs, $X^*$ is the inference point, and $\sigma^2_n$ is the noise level.

Lines 1 and 2 can be computed before we have the inference data. Lines 3 to 5 are for deriving the mean and variance. For the Cholesky decomposition in line 1 and solving the triangular system in lines 2 and 4, the time complexity is $\mathcal{O}(n^3/6)$ and $\mathcal{O}(n^2/2)$, respectively. 

\subsection{Gaussian Process Occupancy Maps (GPOM)}

There are many GPOM frameworks, but most of them only optimize on the Gaussian Process part, and do not make changes in the pipeline, so the framework is still the same. 

The GPOM we will follow is as Algorithm \ref{euclid}. For its input, $\mathbf p$ is the robot pose, $\mathbf r$ is the measurement, $\mathbf \mu$ is the mean, and $\mathbf \sigma$ is the variance. And the output $\mathbf m$ is the occupancy probability.

\begin{algorithm}
	\caption{GPOM for each iteration}\label{euclid}
	\begin{algorithmic}[1]
		\Require $\mathbf p$, $\mathbf r$, $\mu$, $\sigma$
		\State $(\mathbf X, \mathbf y)$ $\leftarrow$ Extract Occupied and free sample  with $\mathbf p$ and $\mathbf r$
		\If {$first\_frame$}
		\State MAP on GP model
		\EndIf
		\State $\mathbf X^*$ $\leftarrow$ Extract inference data with $\mathbf p$
		\State $\mathbf \mu_t^*, \mathbf \sigma_t^* \leftarrow GP(\mathbf X, \mathbf y, \mathbf X^*)$
		\State $\mathbf \mu, \mathbf \sigma \leftarrow BCM(\mathbf \mu, \mathbf \sigma, \mathbf \mu_t^*, \mathbf \sigma_t^*)$
		\State $\mathbf m \leftarrow  \Phi( \frac{+1(\alpha \mathbf \mu + \beta)}{1+\alpha^2 \mathbf \sigma^2})$
		\State \Return $\mathbf m$, $\mathbf \mu$, $\mathbf \sigma$
	\end{algorithmic}
\end{algorithm}

\begin{figure}[]
	\centering
	\subfloat[Positive and Negative samples]{
		\label{fig:clean}
		\includegraphics[width=0.5\linewidth]{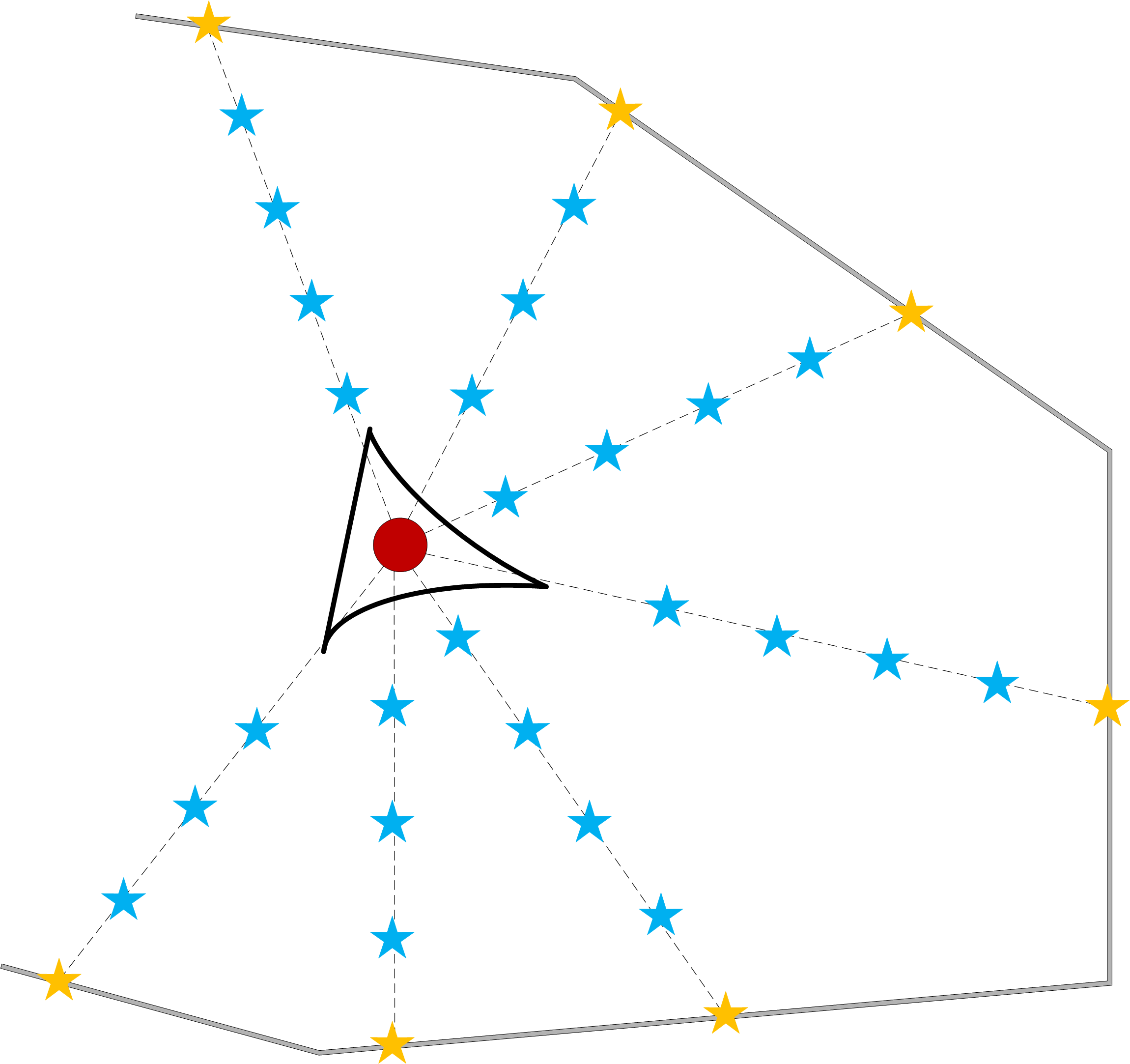}}\\
	\subfloat[Rings]{
		\label{fig:after}
		\includegraphics[width=0.5\linewidth]{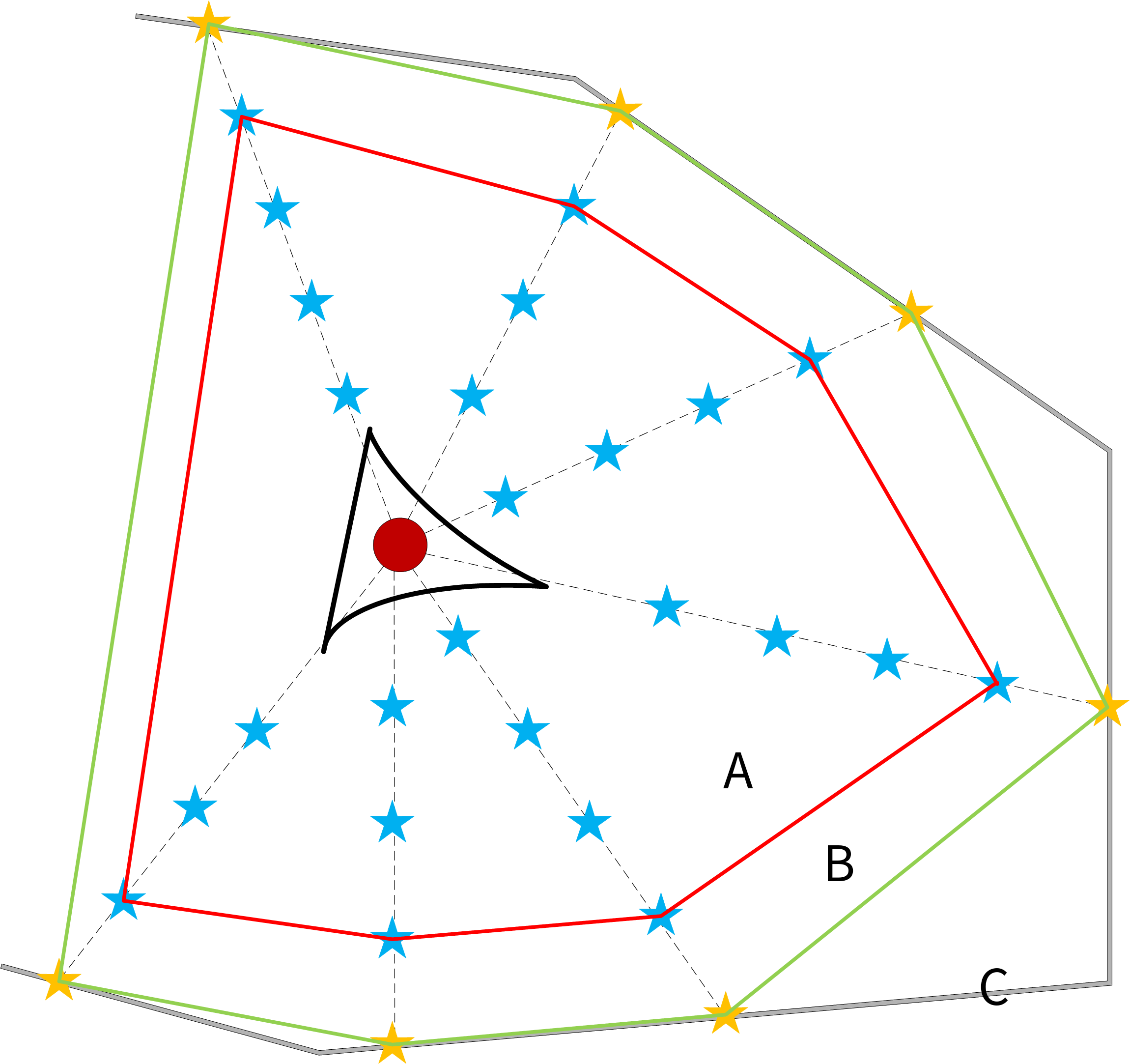}}
	\caption{Extracting samples in GPOM. In the top figure, positive (orange star) and negative (blue star) samples are	extracted from a laser range scan. In the bottom image, the space is separated by two rings: the ring around the obstacles and the ring of the last free sampled points. Thus the space is split into three regions: A, B and C.}
	\label{fig:rings}
\end{figure}

We extract the observed points from the range sensor. From odometry (or an integrated SLAM algorithm) we obtain the robot pose of each frame $\mathbf p_i$. Using the pose $\mathbf p_i$ we can update the map using the information from the range sensor. In Fig. \ref{fig:clean} orange stars represent the scanned obstacle points. By sampling points on the laser line from the robot with an interval $d$ we generate the free samples (blue stars in Fig. \ref{fig:clean}). It should be noted that we don’t add free samples that are closer than $1.5d$ to the occupied point. The second step is to estimate the parameters of the kernel. The first frame is handled specially. Subsequently, we extract the inference data from a window that has the robot pose as the center. This step is just taking many indexes, so it does not take much time. Next, we build the GP model, from \ref{eq:mu}, \ref{eq:sigma}. For that we have to take the inverse of the kernel, which takes $\mathcal{O}(N^3)$. The prediction should take $\mathcal{O}(N^2)$, because it is required to calculate the covariance with training data.

In the end, BCM is used to fuse the prediction and logistic regression to squash the cell values into $[0,1]$.

\section{Our Work}
\label{sec:3}
In our work, we first analyze the computation time for each step of the pipeline and then make optimizations on the identified bottleneck.

\subsection{Pipeline Analysis on GPOM}

The pipeline of GPOM on each iteration is listed as in Section \ref{sec:1}.

The parameter estimation is only for the first frame and will not be included in this analysis. What’s more, we generate the probability map on each frame just for demonstration, it is not really necessary for the GPOM algorithm. Publishing the map in ROS is also extra work that is ignored here.
 
The time cost for all the steps is listed in Table \ref{tab:tab1}, as measured for three different maps. Those maps are shown in Fig. \ref{fig:map_quality}. We can see that there is a conflict with the usual intuition about which step should be optimized. It is not on the GP model building step, which has theoretically a complexity of $O(n^3)$. Instead, the prediction part with a complexity of $O(n^2)$ takes the most of the time. This is not a contradiction, because the two n’s represent different variables.

So let’s be more specific about the computation of GPOM and use an example. We assume a scan with $\#r = 270 $ laser beams (range measurements) and furthermore assume that we extract 5 points on each beam. In experiments, we choose one out of ten lasers to use. So the number of training samples in each iteration should thus be $ n = 27 \times 5 = 135 $. The inference data could be from a window with $ 80 \times 80 $ pixels and then the inference number is $ m = 80 \times 80 = 6400 $.

The computation of the sample extraction is linear with the number of beams $ \#r $, with a small constant weight. The inference data extraction is also linear to the window size. The GP model building and prediction will take $\mathcal{O}(n^3)$ and $\mathcal{O}(m^2)$ while $\mathcal{O}(m^2)$ tends to be much greater. After that, BCM and logistic squash are also linear to $ m $.

The inference computation is comparatively much bigger in applications like GPOM, because it predicts a local map at each frame. This is not the common application scene with GP regression.

\begin{table*}[]
	\centering
		\caption{Time cost on each step in GPOM and Fast-GPOM on three test maps. The time cost of the six steps of the pipeline, the total cost of map building and the cost to publish the map in ROS are demonstrated in this table. The value is an average of the frames after running out of scan data. }
	\begin{tabular}{  |  l ||c |c |c | c| c| c|}
		\hline
		&\multicolumn{3}{|c|}{GPOM cost in milliseconds}&  \multicolumn{3}{|c|}{Fast-GPOM cost in milliseconds} \\ \hline
		& sparse\_obstacles& simple\_rooms & robocup & sparse\_obstacles& simple\_rooms & robocup \\ \hline
		Extract $(X,y)$  &4.63 &4.98 &4.65 &10.57  &5.76  & 65.89\\ \hline
		Extract $X^*$&0.56 &0.21 &0.59  &21.72 &2.23 &11.46\\\hline
		Build GP model&9.45 &7.55 &7.94 &8.02 &3.05 &5.36\\\hline
		Predict&1607.59 &253.34 & 1821.46 &157.97 &8.41 &52.03\\\hline
		BCM&2.12 &0.37 &2.14 &4.68 &0.38 &2.12\\\hline
		Logistic&113.00 &17.58 &113.871 &13.90 &1.27 &7.48 \\\hline \hline
		\textbf{Build Map}&1951.92 &284.31 &1737.20 &219.10 &21.40 &86.02 \\ \hline
		Publish Map &86.87 &14.52 &88.72 &149.65 &14.44 &92.68 \\\hline
	\end{tabular}

	\label{tab:tab1}
\end{table*}
\subsection{Model with Heuristic of spatial information}


Inspired by the above analysis and Table \ref{tab:tab1}, we believe the key to reducing the computation cost is to optimize on the data selection with context-awareness. In this work, we group the space around the robot into three regions.

Spatially, the uncertainty of the laser range sensor has a bound $ d $. It has a high probability to indicate that space away from the obstacle at a distance $ d $ is free. To formulate differently, we can assume that space is free with a small variance. In this paper, we assume that space is free between the robot to $1.5d$ away from the measured obstacle.

We don't observe the space behind the obstacles, so we set the space behind the observed point as unknown. The defect of the classical GPOM is, when it does the model training, in the direction of the laser scan, there should be no observable sample after the occupied sample. Thus we cannot control the value outside the occupied space.

We also assume that if a scan does not provide any useful information on a certain position, the old value should be kept. In Fig. \ref{fig:yellogpom} we can find that for GPOM the space behind the wall has a high probability. But that space has not been observed and thus should not be updated.

First, we first build two rings, as shown in Fig. \ref{fig:rings}. The outer ring is depicted green, it is the polygon border formed with the occupied samples. The inner ring in red is the polygon border formed by the free samples that are closest to the occupied sample on each ray.

The inner and outer rings separate the space into three regions: A (free space with low variance), B (Uncertain region in between the two rings), C (Unknown space).

The strategy we apply for the corresponding regions is:

\begin{enumerate}
	\item Training sample
		\begin{enumerate}
	\item Star points in inner ring as negative samples
	\item Star points in outer ring as positive samples
		\end{enumerate}
	\item Prediction
		\begin{enumerate}
	\item Directly set A region as free space in local frame
	\item Keep the C region value in local frame
	\item Only regression region B
		\end{enumerate}
\end{enumerate}

Then we updated Algorithm \ref{euclid} as Algorithm \ref{ringeuclid}.


\begin{algorithm}
	\caption{Fast-GPOM}\label{ringeuclid}
	\begin{algorithmic}[1]
		\Require $\mathbf p$, $\mathbf r$, $\mu$, $\sigma$
		\State $(\mathbf X, \mathbf y)$ $\leftarrow$ Extract Occupied and free sample  with $\mathbf p$ and $\mathbf r$
		\State  $R_{in}$, $R_{out}$ $\rightarrow$ Extract Inner ring and outer ring
		\State $(\mathbf X^{'}, \mathbf y^{'})$ $\rightarrow$  Selected samples on ring
		\If {$first\_frame$}
		\State MAP on GP model with $(\mathbf X^{'},\mathbf y^{'})$
		\EndIf
		\State $\mathbf X_{in}^*$, $\mathbf X_{mid}^*$, $\mathbf X_{out}^*$ $\rightarrow$ Set inference data in region A, B, and C
		\State $\mathbf \mu_{mid,t}^*, \mathbf \sigma_{mid,t}^* \leftarrow GP(\mathbf X^{'},\mathbf y^{'},\mathbf X_{mid}^*)$
		\State $\mathbf \mu_{in,t}^*, \mathbf \sigma_{in,t}^* \leftarrow \mathbf{-1}, \mathbf{0.1}$
		\State $\mathbf \mu_{out,t}^*, \mathbf \sigma_{out,t}^* \leftarrow \mathbf \mu_{out}, \mathbf \sigma_{out}$		
		\State $\mathbf \mu,\mathbf \sigma \leftarrow BCM(\mathbf \mu,\mathbf \sigma,\mathbf \mu_{t}^*, \mathbf \sigma_{t}^*)$
		\State $\mathbf m \leftarrow  \Phi( \frac{+1(\alpha \mathbf \mu+ \beta)}{1+\alpha^2 \mathbf \sigma^2})$
	\end{algorithmic}
\end{algorithm}

\section{EXPERIMENTS}
\label{sec:4}
To highlight the performance of our method, we compare it with the classical GPOM method in synthetic environments.

\subsection{Experiment Environment}

\begin{figure}[tpb]
	\centering
	\includegraphics[width=0.8\linewidth]{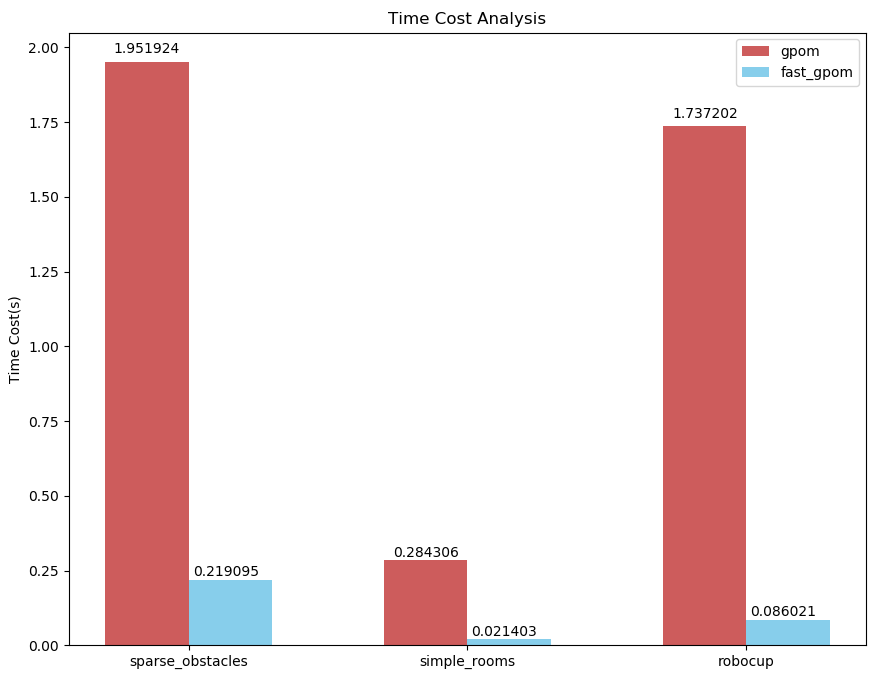}
	\caption{Histogram of the average time cost to build the map for a single scan.}
	\label{fig:time_cost}
\end{figure}

In the experiments, we use a PC with the following configuration: Intel Core i7-6700 3.4GHz and 16GiB memory with Ubuntu 16.04. Our algorithm is implemented based on the Robot Operating System (ROS) and using the Simple Two Dimensional Robot Simulator (STDR) \cite{c9} as our simulation environment. The STDR Simulator is a simple, flexible and scalable robot simulator for use within ROS. It provides a variety of different types of synthetic maps and the ground truth of these maps. It is reasonable to evaluate our algorithm and the performance of the maps in a simulation because then the ground truth is available and we only need to consider the noise of the simulated laser scan data.

In STDR, we utilize a simulated Hokuyo-laser scanner, the simulator robot will provide accurate odometry data. The properties of the simulated Hokuyo laser scanner are the same as real laser scanner and Gaussian noise is applied with a mean, 0.5 and standard deviation, 0.05.

In the experiment, we choose three maps that are provided by the STDR Simulator. The names of the maps are sparse\_obstacles, simple\_rooms and robocup. The sparse\_obstacles map is an indoor scene with several obstacles, the map size is 775 $ \times$ 746 and the resolution of occupancy map is 0.02m. The simple\_rooms map is a basic indoor environment with several empty rooms. This is a relatively simple scenario, and the map size is 600 $\times$ 600 and the resolution of occupancy map is 0.05m. And the robocup map is a hand-draw narrow corridor scenario seem like the environment of robocup rescue league, the map size is 769 $\times$ 729 and the resolution of occupancy map is 0.02. On each map we record  laser scan data and odometry data as a dataset. Then we compare with our algorithm with the general GPOM algorithm using these datasets.

The Gaussian process model is from pyGPs \cite{c10}, which is a widely used library for Gaussian process in python.

For the implementation of the algorithms, GPOM and Fast-GPOM share the same setting, with the same $d$, $W_l$, $H_l$, $W_G$, $H_G$ and Matern (in pyGPs, $d = 7$) kernel. In the squashing step we use $\alpha=100$, $\beta=0$. The resolution of the map is determined to be the same as the arena dataset.

The code, dataset, environment setting and the ROS package of the Fast-GPOM are available on Github\footnote{\url{https://github.com/STAR-Center/fastGPOM}}. 
\subsection{Comparison}
 \newcommand\blaSize{0.23}
\begin{figure*}[]
	\centering
	\subfloat[]{
		\label{fig:exp1:sparse:geo}
		\includegraphics[width=\blaSize\linewidth]{img/experiment1/simple_rooms/gpom_simple_rooms.png}}
	\subfloat[]{
		\label{fig:exp1:sparse:fast}
		\includegraphics[width=\blaSize\linewidth]{img/experiment1/simple_rooms/fast_gpom_simple_rooms.png}}	
	\subfloat[]{
		\label{fig:exp1:sparse:gt}
		\includegraphics[width=\blaSize\linewidth]{img/experiment1/simple_rooms/simple_rooms_ground_truth.png}}
	\subfloat[]{
		\label{fig:exp1:sparse:auc}
		\includegraphics[width=\blaSize\linewidth]{img/experiment1/simple_rooms/simple_rooms_roc_auc.png}}
	\\
	\subfloat[]{
		\label{fig:exp1:simp:geo}
		\includegraphics[width=\blaSize\linewidth]{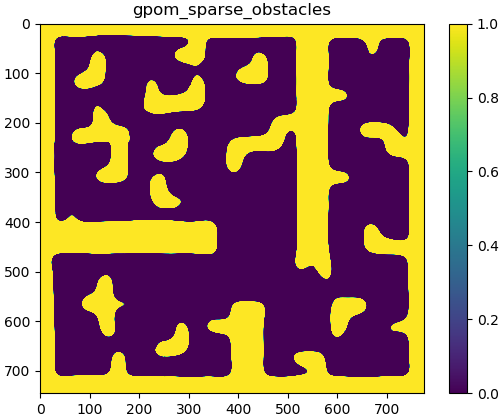}}
	\subfloat[]{
	\label{fig:exp1:simp:fast}
	\includegraphics[width=\blaSize\linewidth]{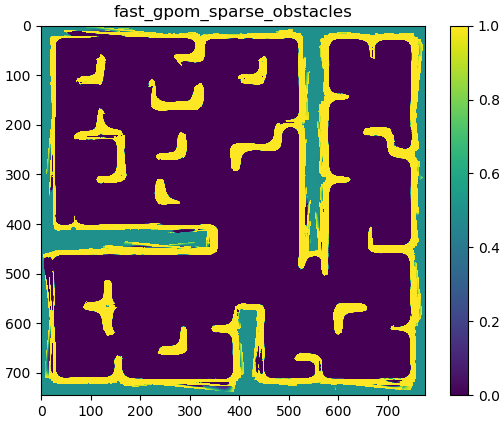}}
	\subfloat[]{
	\label{fig:exp1:simp:gt}
	\includegraphics[width=\blaSize\linewidth]{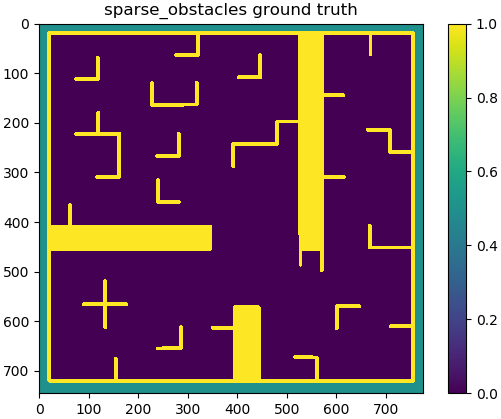}}
	\subfloat[]{
	\label{fig:exp1:simp:auc}
	\includegraphics[width=\blaSize\linewidth]{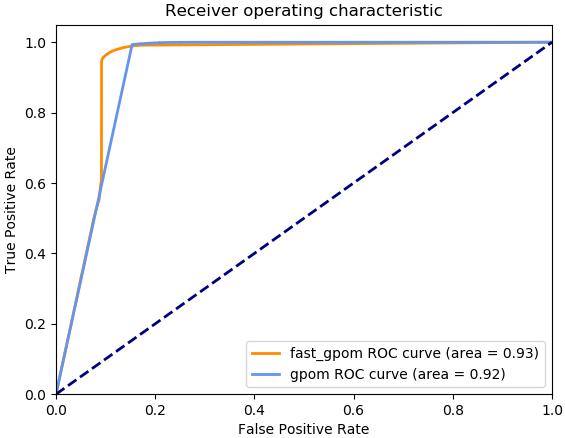}}
	\\
	\subfloat[]{
		\label{fig:exp1:robo:geo}
		\includegraphics[width=\blaSize\linewidth]{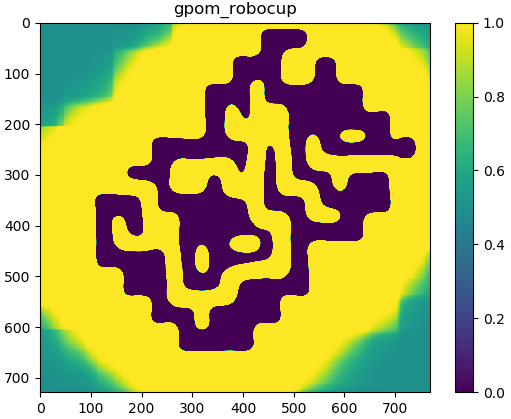}}
	\subfloat[]{
		\label{fig:exp1:robo:fast}
		\includegraphics[width=\blaSize\linewidth]{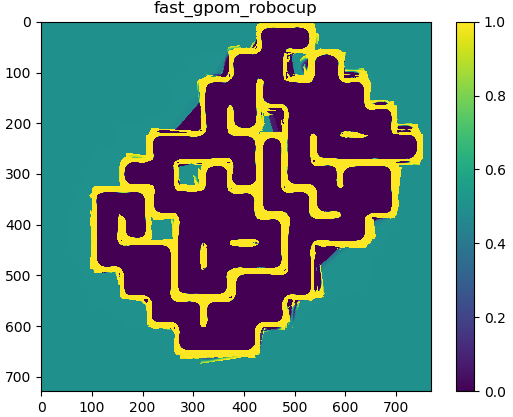}}
	\subfloat[]{
		\label{fig:exp1:robo:gt}
		\includegraphics[width=\blaSize\linewidth]{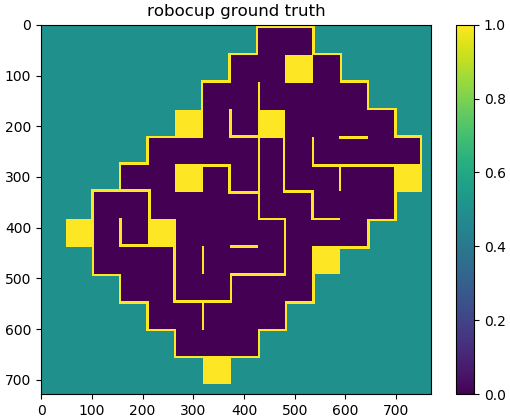}}
	\subfloat[]{
		\label{fig:exp1:robo:auc}
		\includegraphics[width=\blaSize\linewidth]{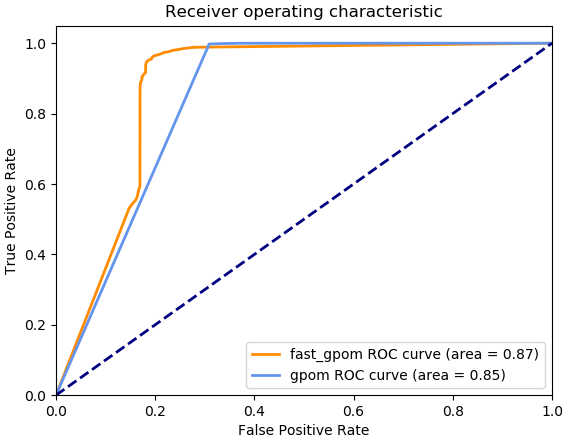}}
	\caption{Map building with GPOM and Fast-GPOM. We show three rows, corresponding to the three maps: simple\_room, sparse\_obstacles and robocup. The first column is for GPOM, the second for our Fast-GPOM. The last two columns are for the ground truth maps and the ROC diagrams, respectively.}
	\label{fig:map_quality}
\end{figure*}

In the experiments we concentrate on three aspects:

\begin{enumerate}
	\item	Time cost analysis on GPOM and Fast-GPOM
	\item	Map quality comparison on GPOM and Fast-GPOM
	\item Map quality comparison on Fast-GPOM with different $d$
\end{enumerate}

In the map quality comparison, we utilize AUC (area under the curve) and ROC (receiver operating characteristic) as metrics to illustrate the performance.

There are different approaches to map quality measurement\cite{c11}. \cite{c5} uses SLAM to estimate the pose and remove some noisy points as the ground truth. In contrast, \cite{c1} only compares the quality to a synthetic map. We follow the latter approach and are using a ground truth map and synthetic sensor data to measure the mapping quality. Our Fast-GPOM is fast enough to work in real time ROS with listening and publishing. But for the experiments, instead of running live, we use recorded data from a bag file, which we read frame by frame.

\subsubsection{Comparison on Time Cost}

The comparison of mapping time cost here is on each simulation arenas. The time cost of each part of the classical GPOM and our Fast-GPOM algorithm are shown in Table \ref{tab:tab1}. For the map building  part, the time cost comparison is illustrated in Fig \ref{fig:time_cost}. We can see that the time cost of our algorithm is reduced tremendously compared to the classical GPOM algorithm. 

From the table, we can see the prediction time of GPOM is in a different order of magnitude compared to the other steps. Thus map building takes much longer than publishing. In contrast, the cost of time in Fast-GPOM is relatively close in among the steps. What’s more, in each arena, the build map cost of Fast-GPOM is in the same magnitude than publishing, in two of those maps it even takes less time. Which means that, if we run map building and publishing in parallel as a ROS package, the map can be displayed smoothly.

The frequency of the simulated laser scan is $10Hz$. In those three maps with the resolution of 0.02m, 0.05m and 0.02m, the frequency to process data can be about $5Hz$, $50Hz$ and $10Hz$. So the Fast-GPOM is adequate for many cases.

\begin{figure}[tpb]
	\centering
	\subfloat[]{
		\includegraphics[width=0.5\linewidth]{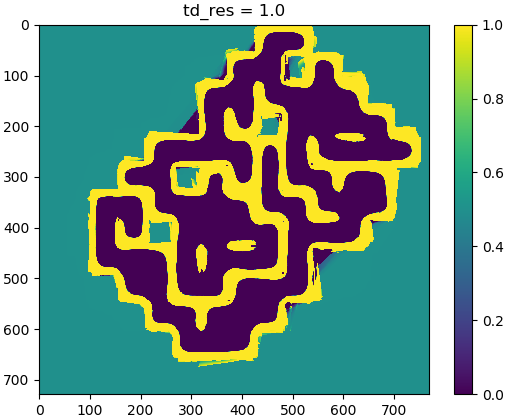}}
	\subfloat[]{
		\label{fig:exp3:sparse:0.5}
		\includegraphics[width=0.5\linewidth]{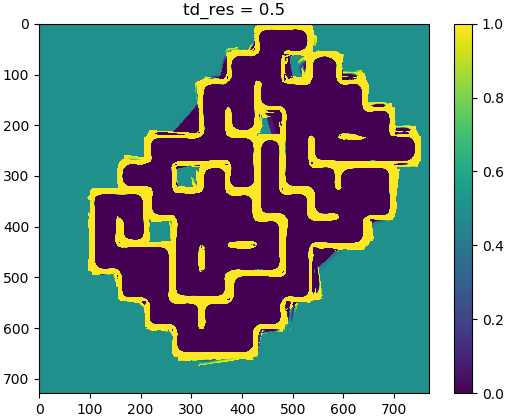}}\\
	\subfloat[]{
		\label{fig:exp3:sparse:0.25}
		\includegraphics[width=0.5\linewidth]{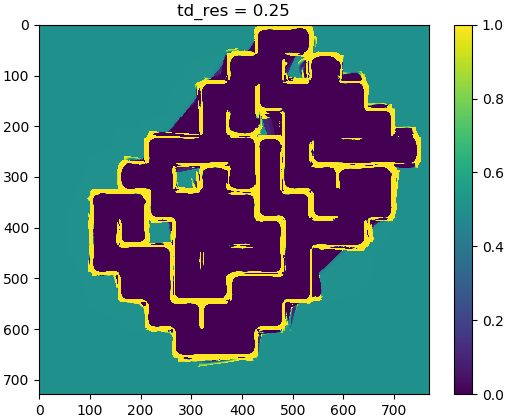}}
	\subfloat[]{
		\label{fig:exp3:sparse:gt}
		\includegraphics[width=0.5\linewidth]{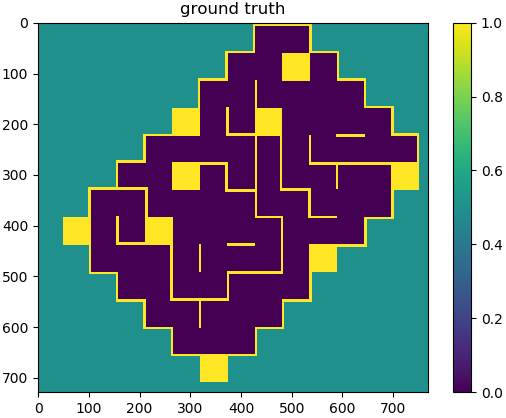}}\\
	\subfloat[]{
		\label{fig:exp3:sparse:auc}
		\includegraphics[width=\linewidth]{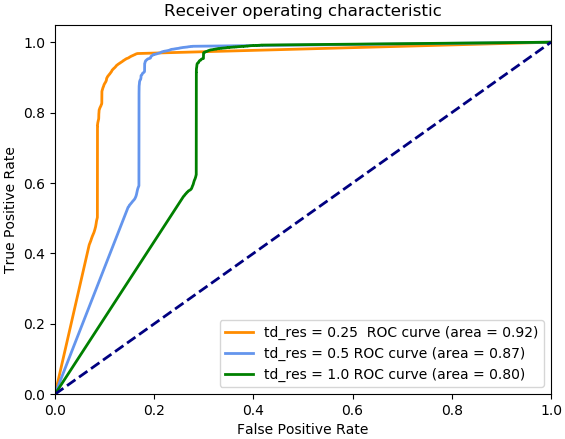}}
	\caption{The generated robocup maps with $ d = 1.0, 0.5, 0.25 $.}
	\label{fig:dist_quality}
\end{figure}

\subsubsection{Comparison of Map quality}
\begin{table}[]
	\centering
	\begin{tabular}{| l | c | c| r |}
		\hline
		&sparse\_obstacles& simple\_rooms & robocup \\ \hline \hline
		GPOM&0.92 &0.95& 0.85\\ \hline
		Fast-GPOM&0.93 &0.93&0.87\\ \hline
	\end{tabular}
	\caption{AUC: GPOM v.s. Fast-GPOM}
	\label{my-auc}
\end{table}
In this part we compare the performance of the mapping between our algorithm and general GPOM by using ROC and AUC. The results of three scenes are showed as Fig \ref{fig:map_quality} and each pixel of the maps is the probability of occupancy. To compute ROC and AUC of each map, we aligned and cropped the resulting map to ensure a pixel to pixel matching with the ground truth. During the evaluation, those pixels that are unknown in ground truth will be neglected. In the Fig. \ref{fig:exp1:sparse:auc}, the AUC of GPOM is higher than our algorithm, because the general GPOM algorithm estimates some parts of the area outside the wall to be occupancy area. That makes the map more like the ground truth, but that is a disadvantage in most scenarios. For example, in Fig. \ref{fig:exp1:robo:geo}, there is a large part of unknown area estimated to be occupied area, but our algorithm could avoid the situations. In the most maps, the AUC of our algorithm is higher than the general GPOM algorithm. That means that our algorithm could guarantee to get a preferable map while it sped up the map building time. The AUC is shown in the ROC figures. To emphasize the comparison, we pick the AUC and put them in Table \ref{my-auc}.

\subsubsection{Map Quality with different $ d $}
In this experiment, we test our Fast-GPOM algorithm with different values for d of the robocup map: $ d = 1.0, 0.5, 0.25 $. We visualize the results and ROC curve in Fig. \ref{fig:dist_quality}.

We can observe that the lower d values tend to provide thinner walls while maybe causing some blur in some edges. The ROC curves indicate that lower d values achieved better AUC value.

\section{CONCLUSIONS}
\label{sec:5}

In this work, we proposed a Fast-GPOM to make the incremental algorithm GPOM much faster. For that, we analyzed the classical Gaussian Process Occupancy Mapping (GPOM) in detail and evaluated the time cost of each step in order to find the most expensive step. Taking the spatial context into consideration, our improvement makes it possible to run GPOM in real time while at the same time reducing the wrong predictions in unknown space. From our experiments, we learned that our method provides a similar or better map quality compared to GPOM while speeding up the computation by an order of magnitude.

\end{document}